\documentclass[times,twocolumn]{article}

\usepackage{framed,multirow}
\usepackage{tabularx}
\usepackage{amssymb}
\usepackage{latexsym}
\usepackage{url}
\usepackage{xcolor}
\definecolor{newcolor}{rgb}{.8,.349,.1}
\usepackage{epsfig}
\usepackage{graphicx}
\usepackage{amsmath}
\usepackage{amsmath}
\usepackage{graphicx}
\usepackage{authblk}
\newcommand{\dt}[2]{\tiny{#1}$\to$\tiny{#2}}
\newcommand{\bata}{JCSL }
\DeclareMathOperator*{\argmin}{argmin}
\DeclareMathOperator*{\argmax}{argmax}

\usepackage{algorithmicx}
\usepackage[ruled]{algorithm}
\usepackage{algpseudocode}

\title{Joint cross-domain classification and subspace learning for unsupervised adaptation}
\date{\vspace{-5ex}}
\begin{document}

\author[1]{Basura Fernando} 
\author[1]{Tatiana Tommasi}
\author[1]{Tinne   Tuytelaars}
\affil[1]{ESAT-PSI/VISICS - iMinds, KU Leuven, Kasteelpark Arenberg 10 , B-3001 Leuven, Belgium}

\maketitle

\begin{abstract}

Domain adaptation aims at adapting the knowledge acquired on a source domain to
a new different but related target domain. Several approaches have been
proposed for classification tasks in the unsupervised scenario, where no labeled target data 
are available. Most of the attention has been dedicated to searching a new domain-invariant 
representation, leaving the definition of the prediction function to a second stage. 
Here we propose to learn both \emph{jointly}. Specifically we learn the source subspace that best 
matches the target subspace while at the same time minimizing a regularized misclassification loss. 
We provide an alternating optimization technique based on stochastic sub-gradient descent 
to solve the learning problem and we demonstrate its performance on several domain 
adaptation tasks.

%
%Unsupervised domain adaptation aims at adapting a classifier trained on a labeled source
%domain to an unlabeled target domain. Towards this goal both the choice of the data representation
%and of the prediction funcion plays an important role, however many existing approaches
%focus only on the first, leaving the classification model to a second stage.

\end{abstract}
\footnotetext[1]{Paper is under consideration at Pattern Recognition Letters.}

\begin{abstract}

Domain adaptation aims at adapting the knowledge acquired on a source domain to
a new different but related target domain. Several approaches have been
proposed for classification tasks in the unsupervised scenario, where no labeled target data 
are available. Most of the attention has been dedicated to searching a new domain-invariant 
representation, leaving the definition of the prediction function to a second stage. 
Here we propose to learn both \emph{jointly}. Specifically we learn the source subspace that best 
matches the target subspace while at the same time minimizing a regularized misclassification loss. 
We provide an alternating optimization technique based on stochastic sub-gradient descent 
to solve the learning problem and we demonstrate its performance on several domain 
adaptation tasks.

%
%Unsupervised domain adaptation aims at adapting a classifier trained on a labeled source
%domain to an unlabeled target domain. Towards this goal both the choice of the data representation
%and of the prediction funcion plays an important role, however many existing approaches
%focus only on the first, leaving the classification model to a second stage.

\end{abstract}

\section{Introduction}

In real world applications, having a probability distribution mismatch between the 
training and the test data is more often the rule than an exception. 
Think about part of speech tagging across different text corpora \cite{Blitzer2006}, 
localization over time with wifi signal distributions that get easily outdated \cite{wifi},
or biological models to be used across different subjects \cite{NIPS2011}. 
Computer vision methods are also particularly challenged in this respect: real world conditions may alter the image 
statistics in many complex ways (lighting, pose, background, motion blur etc.), to not even mention 
the difference in quality of the acquisition device (e.g. resolution), or the high number of possible 
artificial modifications obtained by post-processing (e.g. filtering). 
Due to this large variability, any learning algorithm trained on a \emph{source} set 
regardless of the final \emph{target} data will most likely produce poor, unsatisfactory results.

Domain adaptation techniques propose to overcome these issues and make use 
of information coming from both source and target domains during the learning process.  %to adapt the classifier automatically. 
In the \emph{unsupervised} case, where no labeled samples are provided
for the target, the most extensively studied paradigm consists in assuming 
the existence of a domain-invariant feature space and searching for it. 
In general all the techniques based on this idea focus on transforming the representation of the 
source and target samples to maximize some notion of similarity between them \cite{Gong2012, Gopalan2011,Fernando2013}.
However in this way the classification task is left aside and the prediction model is learned 
only in a second stage. 
As thoroughly discussed in \cite{Ben-David2007,MansourMR09}, the choice 
of the feature representation able to reduce the domain divergence is indeed a crucial factor 
for the adaptation. Nevertheless it is not the only one.
%A common latent space where the difference between the marginal distributions of the two domains 
%is minimized can be identified disregarding the labelling function for the data, which however, 
%is essential to generalize from source to target. 
If several representations induce similar
marginal distributions for the two domains, would a classifier perform equally well in all of them? 
Is it enough to encode the labeling information in the used feature space or is it better to learn 
a cross-domain classification model together with the optimal domain-invariant representation?
Here we answer these questions by focusing on unsupervised domain adaptation subspace solutions.
We present an algorithm that learns jointly both a low dimensional representation and a reliable classifier 
by optimizing a trade-off between the source-target similarity and the source training error. 

%What would happen if we project the source and target data into a low dimensional space that makes 
%the two domains very much alike but loses the discriminative information? As an extreme case, both 
%the domains can be projected to the exact same uninformative point. A common latent subspace where 
%the difference between the marginal distributions of the two domains is minimized can be identified 
%disregarding the conditional distribution of the data, which however, is essential for training a 
%discriminative classifier that generalizes from source to target. If it happens that in several 
%subspaces the domains have similar marginals, would a classifier perform equally well in all of them? 
%Here we answer these questions by %going over the limitations of the existing unsupervised subspace 
%%DA methods and %defining a new algorithm that combines subspace and max-margin learning. We 
%presenting  an algorithm that learns jointly both a low dimensional representation and a reliable classifier 
%by optimizing a trade-off between the source-target similarity and the source training error. 

%

\section{Related Work}
\label{related}

For classification tasks, the goal of domain adaptation is to learn a function from the source domain that 
predicts the class label of a novel test sample from the target domain \cite{patelvisual}. 
In the literature there are two main scenarios depending on the availability of data 
annotations: the semi-supervised and the unsupervised setting.

\vspace{2mm}
In the \textit{semi-supervised} setting a few labeled samples are provided for 
the target domain besides a large amount of annotated source data. 
Existing solutions can be divided into \textit{classifier-based} and \textit{representation-based} methods. 
The former modify the original formulation of Support Vector Machines (SVM) \cite{Yang2007,DuanCVPR2009} 
and other statistical classifiers \cite{DaumeJAIR}: they adapt a pre-trained 
model to the target, or learn the source and target classifiers simultaneously.
The latter exploit the correspondence between source and target labeled data
to impose constraints over the samples through metric learning \cite{Kulis2011,Saenko2010}, 
or consider feature augmentation strategies \cite{Daume07} and manifold alignment \cite{Wang2011}. 
Some approaches have also tackled the cases 
with more than two available domains \cite{multiplesource, Hoffman2012} and the unlabeled 
part of the target has been used for co-regularization \cite{NIPS2010_4009}.
Recently, two methods proposed to combine classifier-based and representation-based solutions. 
\cite{Hoffman2013a} introduced an approach to learn jointly a cross-domain classifier and 
a transformation that maps the target points into the source domain. 
Several kernel maps are used to encode the representation in \cite{DuanTPAMI2012a}, which
proposed a domain transfer multiple kernel learning algorithm. 

\vspace{2mm}
In the more challenging \textit{unsupervised} setting, all the available target samples are 
unlabeled. Many unsupervised domain adaptive approaches resort to estimating the data distributions 
and minimizing a distance measure between them, while re-weighting/selecting the samples \cite{NIPS2011,Gong2013}.
The Maximum Mean Discrepancy (MMD) \cite{Gretton07akernel} maps two sets of data to a reproducing 
Kernel Hilbert Space and it has been largely used as distance measure between two domain distributions. Although 
endowed with nice properties, the choice of the kernel and kernel parameters are critical and, if non-optimal, 
can lead to a very poor estimate of the distribution distance \cite{Gretton:2012}.
Dictionary learning methods have also been used with the goal of defining new representations that overcome the 
domain shift \cite{Ni2013, Shekhar2013}. A reconstruction approach was proposed in \cite{Chang2012}: the
source samples are mapped into an intermediate space where each of them can be represented
as a linear combination of the target domain samples.

Another promising direction for unsupervised domain adaptation is that of \emph{subspace modeling}.
This is based on the idea that source and target share a latent subspace where the domain shift is 
removed or reduced. As for dictionary learning, the approaches presented in this framework are mostly linear, 
but can be easily extended to non-linear spaces through explicit feature mappings \cite{vedaldi11efficient}.
In \cite{Blitzer2011} Canonical Correlation Analysis (CCA) has been applied 
to find a coupled domain-invariant subspace. Principal Component Analysis (PCA) and other eigenvalue
methods are also widely used for subspace generation. For instance, Transfer Component Analysis 
(TCA, \cite{Pan2009}) is a dimensionality reduction approach that searches a latent space 
%spanned by a set of learning components 
where the variance of the data is preserved as much as possible
and the distance between the distributions is reduced. Transfer Subspace Learning (TSL, \cite{Si2010})
couples PCA and other subspace learning methods with a Bregman divergence-based regularization
which measures the distance between the distribution in the projected space. Alternatively, the algorithm 
introduced in \cite{nicta_6981} uses MMD in the subspace to search for a domain invariant projection matrix.
Other methods exploited multiple intermediate subspaces to link the source and the target data. 
This idea was introduced in \cite{Gopalan2011} where the path across the domains is 
defined as a geodesic curve over a Grassmann manifold. This strategy has been further extended in \cite{Gong2012}
where all the intermediate subspaces are integrated to define a cross-domain similarity measure. 
Despite the intuitive characterization of the problem, it is not clear why all the
subspaces along this path should yield meaningful representations. 
Recently the Subspace Alignment (SA) method \cite{Fernando2013} demonstrated that it is possible to map 
directly the source to the target subspace without necessarily passing through intermediate steps.
%or using complicated divergence measures.

\vspace{2mm}
Overall, the main focus of the unsupervised methods proposed in the literature is on the domain invariance of the 
final data representation and less attention has been dedicated to its discriminative power. First
attempts in this direction have been done in \cite{Gong2012} by substituting the use of PCA over the 
source subspace with Partial Least Squares (PLS), and in \cite{Pan2009} where SSTCA chooses the 
representation by maximizing its dependence on the data labels.
Our work fits in this context. \textbf{We aim at extending the integration of classifier-based with 
representation-based solutions in the unsupervised setting where no access
to the target labels is available, not even for hyperparameter cross validation.}
Differently from all the described unsupervised approaches we go beyond searching only a 
domain invariant feature space and we want to optimize also a cross-domain classification model. 
We propose an algorithm that combines effectively subspace and max-margin learning  
and exploits the source discriminative information better than just encoding it
in the representation. Our approach does not need an estimate of the source and target
data distributions and relies on a simple measure of domain shift.
Finally, in previous work the performance of the adaptive methods have been often
evaluated by tuning the model parameters on the target data \cite{Fernando2013,Chang2012}
or by fixing them to default values \cite{Hoffman2013}. Here we choose 
a more fair setup for unsupervised domain adaptation and we show that our approach outperforms different
existing subspace adaptive methods by exploiting exclusively the source annotations.
%Specifically we propose a simple but effective way to combine subspace and max-margin learning.
%, jointly looking for the optimal low-dimensional data projection and 
%for the optimal classifier. 
%PLS is similar to PCA except that it takes label information 
%into consideration and can be seen as a form of supervised dimensionality reduction. 
%Despite its interesting aspect, the combination of PLS with subspace methods is not trivial 
%and leads to unstable results when searching for the optimal subspace dimensinality 
%that both maximizes the source data/label coherence and minimizes the source/target shift.
We name our algorithm \emph{Joint cross-domain Classification and Subspace Learning} (JCSL).

In the following sections we define the notation that will be used in the rest of the paper (section \ref{idea})
and we briefly review the theory of learning from different domains together with the
subspace domain shift measure used in  \cite{Fernando2013} from which we took inspiration. 
We then introduce our approach (section \ref{approach}) followed by an extensive experimental analysis 
that shows its effectiveness on several domain adaptation tasks (section \ref{exper}). We conclude with a 
final discussion and sketching possible directions for future research (section \ref{conclusions}).

\section{Problem Setup and Background}
\label{idea}

Let us consider a classification problem where the data instances are in the form 
$(\mathbf{x}_i,y_i)$. Here $\mathbf{x}_i \in \mathbb{R}^D$ is the feature vector 
for the $i$-th sample and $y_i \in \{1,\ldots,K\}$ is the corresponding label.
We assume that $n_s$ labeled training samples are drawn from a source distribution $\mathcal{D}_s = P(\mathbf{x}^s,y^s)$, 
while a set of $n_t$ unlabeled test samples come from a different target distribution $\mathcal{D}_t = P(\mathbf{x}^t,y^t)$, 
such that it holds $\mathcal{D}_s \neq \mathcal{D}_t$.
In particular, the source and the target distributions satisfy the covariate shift 
property \cite{Shimodaira2000} if they have the same labeling function with 
$P(y^s|\mathbf{x}^s)=P(y^t|\mathbf{x}^t)$, while the marginal distributions differ 
$P(\mathbf{x}^s)\neq P(\mathbf{x}^t)$. We operate under this hypothesis.
%In this case the domain difference reduces to a marginal distribution shift.

\paragraph{A bound on the target domain error} 
Theoretical studies on domain adaptation have established the conditions under which a classifier trained 
on the source data can be expected to perform well on the target data. 
The following generalization bound on the target error $\epsilon_t$ has been 
demonstrated in \cite{Ben-David2007}:
\begin{equation}
 \epsilon_t(h) \leq \epsilon_s(h) + d_{\mathcal{H}}(\mathcal{D}_s,\mathcal{D}_t) + \lambda~.
 \label{bendavid}
\end{equation}
Here $h$ indicates the predictor function, while $\mathcal{H}$
is the hypothesis class from which the predictor has been chosen. In words, the bound states that
a low target error can be guaranteed if the source error $\epsilon_s(h)$, a measure of the domain 
distribution divergence $d_{\mathcal{H}}(\mathcal{D}_s,\mathcal{D}_t)$, and the error $\lambda$ of 
the ideal joint hypothesis on the two domains are small. 
The joint error can be written as \mbox{$\lambda = \epsilon_t(h^*) + \epsilon_s(h^*)$}
where \mbox{$h^* = \argmin_{h \in \mathcal{H}} (\epsilon_t(h) + \epsilon_s(h))$}. The value
of $\lambda$ is supposed to be low under the the covariate shift assumption.

%When this hypothesis performs poorly, we cannot expect to learn a good target classifier in an 
% unsupervised domain adaptation setting where only the source error can be estimated, even though 
% the feature representation has been chosen to minimize the divergence of the marginal distributions. 
%All the three terms on the right hand side of (\ref{bendavid}) contribute to the adaptability
%of the classifier. 

\paragraph{A subspace measure of domain shift}
The low-dimensional intrinsic structure of the source and target domains can be specified 
by their corresponding orthonormal basis sets, indicated respectively as $S \in \mathbb{R}^{D\times d}$ and 
$T \in \mathbb{R}^{D\times d}$. These are two full rank matrices, and $d$ is the subspace dimensionality. 
In \cite{Fernando2013}, a transformation matrix $M$ is introduced to modify the source subspace.
The domain shift of the transformed source basis with respect to the target is simply measured by the 
following function:
\begin{equation}
F(M) = || S M - T ||_F^{2} ~, 
\label{eq:objective-sa}
\end{equation}
where $||\cdotp||_F$ is the Frobenius norm. The \emph{Subspace Alignment} (SA) method proposed to 
minimize this measure, obtaining the optimal transformation matrix in closed form: $M=S^\top T \in \mathbb{R}^{d \times d}$.
The matrix $U=SM=SS^\top T \in\mathbb{R}^{D \times d}$ is finally used to represent the source data.
The original domain basis sets can be obtained through different strategies, 
both unsupervised (PCA) and supervised (PLS, LDA), as extensively studied in \cite{Fernando2014}.
%With PCA the basis correspond to the top $d$ eigenvectors of the data's covariance matrix.
%Alternatively, discriminative subspaces can be determined by using the source labels with 
%PLS or Linear Discriminant Analysis (LDA). 
%In this last case the resulting subspace dimensionality 
%depends on the number of classes by $d=C-1$.
%such that the distribution of the source reprojected data $(\mathbf{x}^sU)$ matches
%the distribution of the subspace target samples $(\mathbf{x}^tT)$.

SA has shown promising results for visual cross-domain 
classification tasks outperforming other subspace adaptive methods.
However, on par with its competitors \cite{Gopalan2011,Gong2012}, it keeps the domain 
adaptive process (learning M) and the classification process (e.g. learning an SVM model) 
separated, focusing only on the distribution divergence term $d_{\mathcal{H}}(\mathcal{D}_s,\mathcal{D}_t)$
of the bound in (\ref{bendavid}).

\section{Proposed Approach}
\label{approach}

With the aim of minimizing both the domain divergence and the source error in (\ref{bendavid}), we
propose an algorithm that learns a domain-invariant representation and an optimal cross-domain 
classification model.
%When focusing on covariate shift problems (negligeble $\lambda$), the target performance can be optimized 
%by considering the last two terms in the right hand side of the bound in (\ref{bendavid}).
%The goal is to search jointly for a representation that minimizes the domain divergence and for 
%the best classification model that minimizes the source error in the same space.
For the representation we concentrate on subspace methods and we take inspiration from the SA approach.
For the classification we rely on a standard max-margin formulation. 
The details of our \emph{Joint cross-domain Classification and Subspace Learning} (JCSL) algorithm are 
described below.

Given a fixed target subspace basis $T \in \mathbb{R}^{D \times d}$ we minimize the following 
regularized risk functional 
\begin{equation}
G(V,\mathbf{w}) =  ||\mathbf{w}||_2^2+  \alpha || V - T ||_F^2 + \beta\sum_i^{n_s} \mathcal{L}(\mathbf{x_i^s},y_i^s,\mathbf{w},V)~.
\label{riskfunctional}
\end{equation}
Here the regularization terms aim at optimizing separately the linear source classification model $\mathbf{w} \in \mathbb{R}^d$,
and the source representation matrix $V\in \mathbb{R}^{D \times d}$, while the loss function $\mathcal{L}$
depends on their combination. 
For our analysis we choose the hinge loss: 
$\mathcal{L}(\mathbf{x_i^s},y_i^s,\mathbf{w},V)=\max\{0,1 - \mathbf{x_i^s}^\top V \mathbf{w} y_i^s\}$,
but other loss functions can be used for different cross-domain applications.
The parameters $\alpha$ and $\beta$ allows to define a trade-off between the importance of the 
terms in the objective function. In particular a high $\alpha$ value pushes $V$ towards $T$ giving
more importance to the distribution divergence term, while a high $\beta$ value focuses the attention 
on the training error term to improve the classification performance in the new space.

The matrix $V$ has a role analogous to that of $U$ in SA, however 
in our case it is not necessary to specify a priori the source subspace $S$ which is now optimized together with the 
alignment transformation matrix $M$ in a single step. Note that, if the source and target data can be considered as 
belonging to the same domain (no domain shift), our method will automatically provide $V=T$ boiling down to standard 
learning in the shared subspace.
%This gives to JCSL the possibility to be used even for asymmetric cross domain problems where the
%source and target data are represented by different dimensional features.
We follow previous literature and propose the use of PCA to define the target subspace $T$ \cite{Fernando2013,Gong2012}.
Besides having demonstrated good results empirically, the theoretical meaning of this choice can
be identified by writing the mutual information between the target and the source as
\begin{equation}
MI(source;target) = H(target) - KL(source||target)~.
\end{equation}
%$MI(U;T) = H(T) - KL(U||T)$
%where $KL(\cdotp||\cdotp)$ is the Kullback-Leibler divergence and $H(\cdotp)$ is the entropy. 
Projecting the target data to the subspace $T$ maximizes the entropy $H(target)$, 
while our objective function minimizes the domain shift, which is related to the 
Kullback-Leibler divergence $KL(\cdotp||\cdotp)$. % by the regularization term on $U$. 
Hence, we expect to increase the mutual information between source and target.
%A second reason for our choice comes directly from the empirical good results already shown in
%previous works that were adopting PCA for subspace adaptive methods \cite{Fernando2013,Gong2012}.

Minimizing (\ref{riskfunctional}) jointly over $(V,\mathbf{w})$ is a non-convex problem and finding
a global optimum is generally intractable. However we can apply alternated minimization for $V$ 
and $\mathbf{w}$ resulting in two interconnected convex problems that can be efficiently solved by 
stochastic subgradient descent. For this procedure we need the partial derivatives of 
(\ref{riskfunctional}) that can be easily calculated as:
\begin{equation}
{\arraycolsep=1.4pt\def\arraystretch{2.2}
\begin{array}{ll}
 \dfrac{\partial G(V,\mathbf{w})}{\partial V} =& 2\alpha( V - T ) -\beta ~\Sigma_{i=1}^{n_s} ~\Gamma_i~\\
 \dfrac{\partial G(V,\mathbf{w})}{\partial \mathbf{w}} =& 2\mathbf{w} -\beta~ \Sigma_{i=1}^{n_s} ~\Theta_i~
\end{array}
}
\vspace{1mm}
\end{equation}
where  $\Gamma$ and $\Theta$ are the derivatives of $\mathcal{L}(\mathbf{x_i^s},y_i^s,\mathbf{w},V)$
with respect to $V$ and $\mathbf{w}$. When using the hinge loss we get

\vspace{3mm}
\begin{tabular}{ccc}
%\begin{equation} 
$
\Gamma_i = \left\{
%{\arraycolsep=1.4pt\def\arraystretch{1.5}
%\begin{array}{lll}
% \mathbf{x_i^s}^\top \mathbf{w} y_i^s~,  &~~~ \Theta_i= \mathbf{x_i^s}^\top V y_i^s  &~~~\text{if}~(\mathbf{x_i}^\top V \mathbf{w} y_i) < 1 \\
% 0~,                                     &~~~ \Theta_i=0			       &~~~~\text{otherwhise}~.
%\end{array}
%}\vspace{1mm}
 \begin{array}{l}
  \mathbf{x_i^s}^\top \mathbf{w} y_i^s \\
  0 
  \end{array}
\right.
$%\end{equation}
&
$%\begin{equation}
\Theta_i = \left\{
  \begin{array}{ll}
  \mathbf{x_i^s}^\top V y_i^s\\
  0
  \end{array}
\right.
$%\end{equation}
& 
$%\begin{equation}
 \begin{array}{l}
\text{if}~(\mathbf{x_i^s}^\top V \mathbf{w} y_i^s) < 1 \\
\text{otherwise}~. \nonumber 
  \end{array}
$%\end{equation}
\end{tabular}
%\end{equation}
\vspace{1mm}

The iterative subgradient descent procedure terminates when the algorithm converges, 
showing a negligible change of either $V$ or $\mathbf{w}$ between two consecutive iterations.
The formulation holds for a binary classifier but can easily be  used in its one-vs-all multiclass
extension that highly benefits from the choice of the stochastic variant of the optimization process.

At test time, we indicate the classification score of class $y$ for the target sample  
$\mathbf{x_i^t}$ as $s(\mathbf{x_i^t}, \mathbf{w}_y)=\mathbf{x_i^t}^\top T \mathbf{w}_y$ .
The multiclass final prediction is then obtained by maximizing over the scores: 
$y_i^* = \argmax_y(s(\mathbf{x_i^t}, \mathbf{w}_y))$.
Note that the source representation matrix $V$ is not involved at this stage, and the 
target subspace basis $T$ appears instead. Differently from the pre-existing unsupervised 
domain adaptation methods that encode the discriminative information in the representation,
JCSL learns directly a domain invariant classification model able to generalize from source
to target.
The JCSL learning strategy is summarized in Algorithm \ref{ALG}.

%\vspace{-0.2cm}%
\alglanguage{pseudocode}
\begin{algorithm}[h!]
\small
\caption{\bata}
\label{ALG}
\begin	{algorithmic}[1]
\Statex \textbf{Input:} step size $\eta$ and batch size $\gamma$ for stochastic sub-gradient descent 
                        %$(\mathbf{x_i^s}, y_i^s)_{i=1}^{n_s}$, $T$, $\alpha$, $\beta$, $d$.
\Statex \textbf{Output:} $V^*$,$\mathbf{w}^*$
\State Initialize $V \leftarrow S$,$\mathbf{w}\leftarrow 0$, $k\leftarrow0$
    \While {not converged}
      \State $k\leftarrow k+1$
      \State calculate the partial derivatives:
      \vspace{2mm}
      \Statex $~~~~~~\frac{\partial G(V,\mathbf{w})}{\partial V} = 2\alpha( V - T ) -\beta \Sigma_{i=1}^\gamma \Gamma_i$
      \vspace{2mm}
      \Statex $~~~~~~\text{with}~~\Gamma_i = \left\{
      \begin{array}{ll}
      \mathbf{x_i^s}^\top \mathbf{w} y_i^s   & ~~  \text{if}~~ (\mathbf{x_i^s}^{\top} V \mathbf{w} y_i^s) < 1\\
      0  & ~~ \mbox{otherwise}\\
      \end{array}
      \right.$
      \vspace{2mm}
      \Statex $~~~~~~\frac{\partial G(V,\mathbf{w})}{\partial \mathbf{w}} = 2\mathbf{w} -\beta \Sigma_{i=1}^\gamma \Theta_i$
      \vspace{2mm}
      \Statex $~~~~~~\text{with}~~\Theta_i = \left\{
     \begin{array}{ll}
     \mathbf{x_i^s}^\top V y_i^s   & ~~  \text{if}~~ (\mathbf{x_i^s}^\top V \mathbf{w} y_i^s) < 1\\
     0  & ~~ \mbox{otherwise}\\
     \end{array}
     \right.$
     \vspace{2mm}
     \State  Fix $V$, identify the optimal $\mathbf{w}$: 
      \vspace{2mm}
      \Statex  $~~~~~~\mathbf{w}_k \leftarrow \mathbf{w}_{k-1} - \eta \left(\frac{\partial F(V,\mathbf{w})}{\partial \mathbf{w}}\right)_{\mathbf{w}_{k-1}}$
      \vspace{2mm}
      \State  Fix $\mathbf{w}$, identify the optimal $V$: 
      \vspace{2mm}
      \Statex $~~~~~~V_k \leftarrow V_{k-1} - \eta\left(\frac{\partial F(V,\mathbf{w})}{\partial V}\right)_{V_{k-1}}$
    \EndWhile
\Statex
\end{algorithmic}
  \vspace{-0.4cm}%
\end{algorithm}

\vspace{-3mm}
\section{Experiments}
\label{exper}

We validate our approach over several domain adaptation tasks. In the following
we first describe our experimental setting (section \ref{exp:setup}) and then we report on the 
obtained results (sections \ref{sec:results}, \ref{sec:resultsBING}, \ref{sec:resultsWIFI}). 
Moreover we present a detailed analysis on the role of the learning parameters and on the 
domain-shift reduction effect of \bata (section \ref{sec:results}).
%with a detailed analysis on how \bata effectively optimizes over the source classification 
%error and the cross-domain similarity.

\begin{figure*}[t]
\centering
 \begin{tabular}{|c|} \hline
\includegraphics[width=0.95\textwidth]{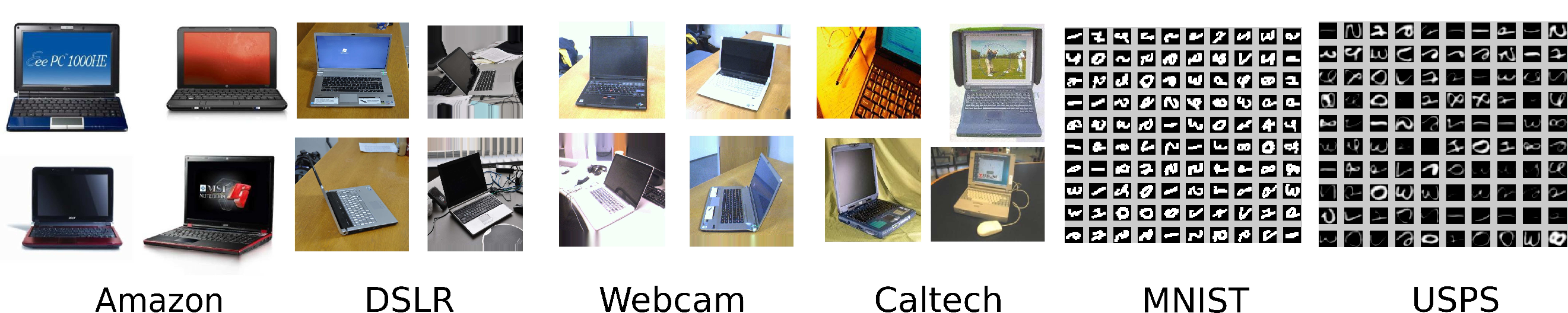}\\\hline
\includegraphics[width=0.9\textwidth]{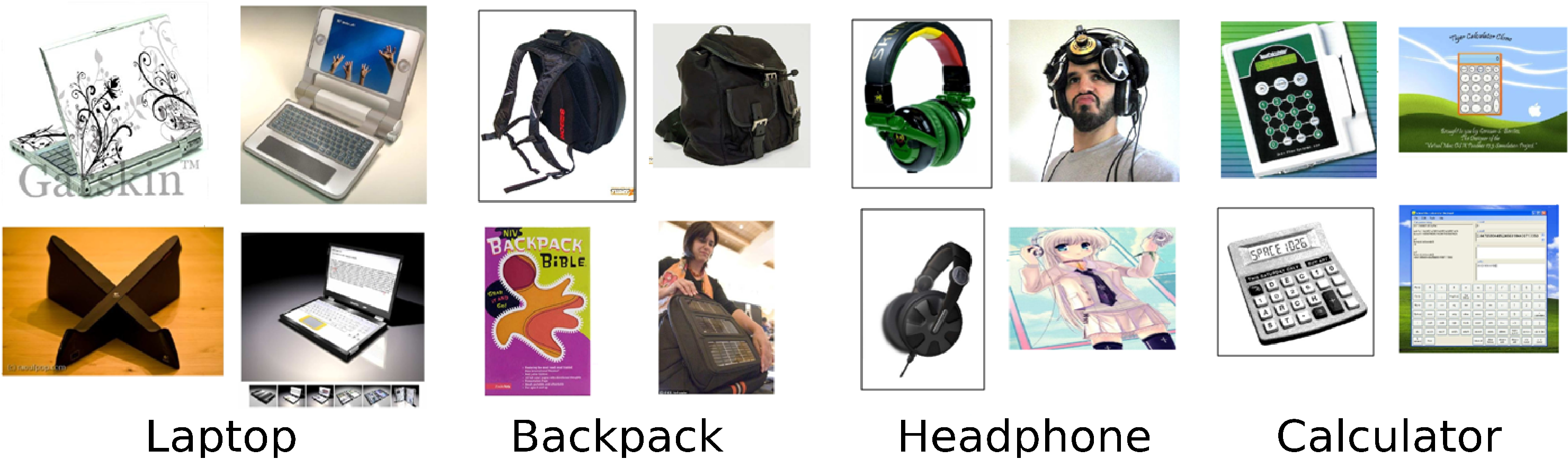}\\ \hline
 \end{tabular}
\caption{Top line: examples from Office+Caltech dataset and the MNIST+USPS 
dataset. Bottom line: weakly labeled images from Bing dataset.}
 \label{fig:datasets}
\end{figure*}  
\begin{table*}[t]
\caption{Recognition rate (\%) results over the Office+Caltech and MNIST+USPS datasets.}
\centering
\scriptsize
\begin{tabular}{ c c c c c c c c c }\hline
DA Problem 	& NA 	 		& $PCA_T$	& SA$_{(LDA-PCA)}$         & GFK$_{(LDA-PCA)}$	& TCA                    	& SSTCA			& TSL        		& \bata \\ \hline 
\dt{A}{C}	&38.1 $\pm$ 2.6 	&41.1$\pm$1.7	&43.4 $\pm$ 3.2 	&43.2 $\pm$ 3.7 	&\textbf{43.5 $\pm$ 3.2 }	&38.8$\pm$2.4		&40.4$\pm$0.9		&42.6 $\pm$ 0.9  \\ 	
\dt{A}{D}	&32.9 $\pm$ 2.8 	&37.5$\pm$1.4	&\textbf{44.7 $\pm$ 2.6} &43.7 $\pm$ 2.8 	&38.8 $\pm$ 2.1 		&34.1$\pm$6.9		&40.8$\pm$1.7		&42.5 $\pm$ 3.2  \\ 	
\dt{A}{W}	&36.8 $\pm$ 2.9 	&39.1$\pm$2.8	&40.3 $\pm$ 2.9 	&41.3 $\pm$ 2.0 	&41.0 $\pm$ 1.0 	  	&34.1$\pm$3.8		&41.1$\pm$2.3		&\textbf{47.6 $\pm$ 2.1}  \\ 	
\dt{C}{A}	&39.5 $\pm$ 1.0 	&40.6$\pm$2.8	&39.3 $\pm$ 1.8 	&39.9 $\pm$ 1.5 	&42.5 $\pm$ 1.2 	  	&39.1$\pm$3.5		&43.0$\pm$4.2		&\textbf{44.3 $\pm$ 1.2}  \\ 	
\dt{C}{D}	&38.8 $\pm$ 2.4 	&40.9$\pm$1.3	&44.0 $\pm$ 2.0 	&42.0 $\pm$ 2.5 	&42.1 $\pm$ 2.5 	  	&38.3$\pm$4.1		&40.4$\pm$3.4		&\textbf{46.5 $\pm$ 1.5}  \\ 	
\dt{C}{W}	&37.8 $\pm$ 1.4 	&35.9$\pm$3.2	&37.3 $\pm$ 3.3 	&41.8 $\pm$ 3.8 	&39.4 $\pm$ 2.5 	  	&31.6$\pm$4.9		&40.9$\pm$2.4		&\textbf{46.5 $\pm$ 2.0}  \\ 	
\dt{D}{A}	&24.4 $\pm$ 1.7 	&30.6$\pm$2.7	&35.7 $\pm$ 2.3 	&31.0 $\pm$ 2.7 	&30.4 $\pm$ 2.1 	  	&38.0$\pm$2.6		&39.6$\pm$1.2		&\textbf{41.3 $\pm$ 0.9}  \\ 	
\dt{D}{C}	&30.5 $\pm$ 2.1 	&37.9$\pm$1.3	&\textbf{41.5 $\pm$ 1.4} &40.9 $\pm$ 2.8 	&36.7 $\pm$ 2.6 		&32.9$\pm$1.8		&33.4$\pm$2.1		&35.1 $\pm$ 0.9  \\ 	
\dt{D}{W}	&60.9 $\pm$ 3.1 	&67.9$\pm$2.8	&58.6 $\pm$ 2.1 	&60.5 $\pm$ 3.8 	&64.5 $\pm$ 3.2 	  	&\textbf{76.2$\pm$3.0}	&73.7$\pm$1.5		&74.2 $\pm$ 3.6  \\ 	
\dt{W}{A}	&29.7 $\pm$ 1.7 	&34.1$\pm$1.7	&34.7 $\pm$ 0.9 	&33.1 $\pm$ 1.2 	&34.6 $\pm$ 1.4 	  	&35.1$\pm$2.6		&38.0$\pm$1.6		&\textbf{43.1 $\pm$ 1.0}  \\ 	
\dt{W}{C}	&34.1 $\pm$ 1.6 	&38.1$\pm$1.3	&36.7 $\pm$ 1.2 	&37.7 $\pm$ 1.8 	&\textbf{39.6 $\pm$ 2.2} 	&29.7$\pm$2.5		&30.4$\pm$1.3		&36.1 $\pm$ 2.0  \\ 	
\dt{W}{D}	&71.1 $\pm$ 2.6 	&74.6$\pm$0.9	&69.5 $\pm$ 2.4 	&75.3 $\pm$ 2.6 	&\textbf{77.3 $\pm$ 2.7} 	&69.9$\pm$3.4		&66.9$\pm$1.6		&66.2 $\pm$ 2.9  \\ \hline 
AVG. 		&39.6 		 	&43.2		&43.8		 	&44.2			&44.2				& 41.4	 		& 44.0			&\textbf{47.2}  \\ \hline\hline 
\dt{MNIST}{USPS}& 45.4 			&45.1 	        &\textbf{48.6}		& 34.6 			&40.8				& 40.6			& 43.5 			& 46.7 \\ 	
\dt{USPS}{MNIST}& 33.3 			&33.4 		& 22.2 			& 22.6 			&27.4				& 22.2			& 34.1 			& \textbf{35.5} \\ \hline 
AVG. 		& 39.4 			&39.2		& 35.4 			& 28.6 			&34.1				& 31.4			& 38.8 			& \textbf{41.1}\\ \hline 
\end{tabular}
\label{tbl:results}
\end{table*}

\subsection{Datasets, baselines and implementation details}
\label{exp:setup}

We choose three image datasets (see Figure \ref{fig:datasets}) and a wifi signal dataset.

\vspace{2mm}
\textbf{Office + Caltech \cite{Gong2012}}. This dataset was created  by combining the 
Office dataset \cite{Saenko2010} with Caltech256 \cite{Griffin2007} and it contains 
images of 10 object classes over four domains: \textbf{A}mazon, \textbf{D}slr, \textbf{W}ebcam and \textbf{C}altech.
Amazon consists of images from online merchants' catalogues, while 
Dslr and Webcam domains are composed by respectively high and low resolution images.
Finally, Caltech corresponds to a subset of the original Caltech256. We use the features
provided by Gong et al. \cite{Gong2012}:
SURF descriptors quantized into histograms of 800 bag-of-visual words and standardized by z-score normalization.
All the 12 possible source-target domain pairs are considered. We use the data splits provided by Hoffman et al. \cite{Hoffman2013a}.

\vspace{2mm}
\textbf{MNIST \cite{LeCun1998} + USPS \cite{Hull1994}}. This dataset combines two existing image
collections of digits presenting different gray scale data distributions. Specifically they share 10 classes 
of digits. We randomly selected 1800 images from USPS and 2000 images from MNIST. By following 
\cite{Long2013} we uniformly re-scale all images to size 16 $\times$ 16 and we use the L2-normalized gray-scale pixel 
values as feature vectors. Both domains are alternatively used as source and target.

\vspace{2mm}
\textbf{Bing+Caltech \cite{Bergamo2010}}. In this dataset, weakly annotated images from the Bing search engine 
define the source domain while images of Caltech256 are used as target. We run experiments varying the
number of categories (5, 10, 15, 20, 25 and 30) and the number of source examples per category (5 and 10) using the same
train/test split adopted in \cite{Bergamo2010}. As typically done for this dataset, Classemes features are used as 
image representation \cite{Bergamo2010}.
 
\vspace{2mm}
\textbf{WiFi \cite{wifi}}. This dataset was used in the 2007 IEEE ICDM contest for domain adaptation.
The goal is to estimate the location of mobile devices based on the received signal strength (RSS) values
from different access points. The domains correspond to two different time periods during which the 
collected RSS values present different distributions. The dataset contains 621 labeled examples collected
during time period A (source) and 3128 unlabeled examples collected during time period B (target). 
The location recognition performance is generally evaluated by measuring the average error distance
between the predicted and the correct space position of the mobile devices. We slightly modify the task
to define a classification rather than a regression problem. We consider 247 locations and we evaluate 
the classification performance between the sets A and B with and without domain adaptation.
%There are 248 locations in total, each considered as a class. 
%All source samples are used during the training and the classification accuracy is measured using all 
%test samples.
We repeat the experiments both testing over all the target data and considering 10
random target splits, each with 400 random samples.

\vspace{2mm}
We benchmark \bata\footnote{We implemented our algorithm in MATLAB. The code is submitted with the 
paper as supplementary material.} against the following subspace-based domain adaptation methods:
%\begin{description}

\vspace{2mm}
\textbf{TCA, SSTCA}: Transfer Component Analysis and its semi-supervised extension \cite{Pan2009}.
We implemented TCA and SSTCA by following the original paper description. For SSTCA 
we turned off the locality preserving option to have a fair comparison with all the other 
considered methods, none of which exploits local geometry\footnote{Following the original
paper notation we fixed $\mu=0.1$, $\lambda=0$, $\gamma=0.5$.}. 
%The objective function 
%optimizes a transformation matrix and presents two terms, a regularizer to control the
%complexity of the transformation and a measure of the domain distance in the new space. 
%We fixed the trade-off parameter among these two terms to $0.1$ (parameter $\mu$ in 
%the notation of \cite{Pan2009}). We used the same parameter value for SSTCA and we turned off
%the locality preserving option to have a fair comparison with all the other benchmarked
%methods, none of which exploits local geometry. Furthermore we followed the instructions in the paper to set the SSTCA balance 
%parameter between the label dependence and the data variance term ($\lambda=0.5$ according
%to \cite{Pan2009} notation).

\vspace{2mm}
\textbf{TSL}: Transfer Subspace Learning \cite{Si2010}.
We used the code made publicly available by the authors\footnote{\url{http://www.cs.utexas.edu/~ssi/TrFLDA.tar.gz}} 
which implements TLS by adding a Bregman-divergence based regularization to the Fisher's Linear Discriminant
Analysis (FLDA). 
%FLDA exploits the source labels and maximizes the mean value 
%of Kullback-Leibler divergences between the different classes in the hypothesis that 
%they are sampled from Gaussian distributions with different means but identical covariance.

\vspace{2mm}
\textbf{GFK}$_{\mathbf{(LDA-PCA)}}$, \textbf{SA}$_{\mathbf{(LDA-PCA)}}$: for both the Geodesic Flow Kernel \cite{Gong2012} and the 
Subspace Alignment \cite{Fernando2013} methods we slightly modified the original implementation 
provided by the authors\footnote{
\url{http://www-scf.usc.edu/~boqinggo/domain_adaptation/GFK_v1.zip}, 
\url{http://homes.esat.kuleuven.be/~bfernand/DA_SA/downloadit.php?fn=DA_SA.zip}}
to integrate the available discrimintive information in the source domain. 
%Specifically we used LDA instead of PCA to create the source subspace.
%\textbf{GFK}: Geodesic Flow Kernel \cite{Gong2012}. We use the GFK implementation provided by
%the authors\footnote{\url{http://www-scf.usc.edu/~boqinggo/domain_adaptation/GFK_v1.zip}} and define the source subspace through LDA.
%
%\vspace{2mm}
%\textbf{SA}:  Subspace Alignment \cite{Fernando2013}. We use the SA implementation provided by
%the authors\footnote{\url{http://homes.esat.kuleuven.be/~bfernand/DA_SA/downloadit.php?fn=DA_SA.zip}} and define the source subspace through LDA.
%%\end{description}
%\vspace{2mm}
As preliminary evaluation we compared the results of GFK and SA when the basis of the 
source subspace were obtained with PLS and LDA. Although performing similarly on average, 
PLS showed less stability than LDA with large changes in the outcome for small variations of the
subspace dimensionality $d$. This can be explained by considering the difficulty of finding the 
best $d$ that jointly maximizes the source data/label coherence and minimizes the source/target shift.
Thus, for our experiments we rely on the more stable LDA for the source which fixes $d$ equal $K-1$.
%Despite its interesting aspect, the combination of PLS with subspace methods is not trivial 
%and leads to unstable results when searching for the optimal subspace dimensinality 
%that both maximizes the source data/label coherence and minimizes the source/target shift.
%From a preliminary evaluation GFK and SA have shown their best performance when LDA is used 
%to generate a discriminative source subspace.
On the other hand, the target subspace is always obtained by applying PCA and selecting the 
first $K-1$ eigenvectors. 

\vspace{2mm}
As further baselines we also consider the source classifier learned with no adaptation (\textbf{NA}) in the 
original feature space  and in the target subspace. The last one is obtained by
applying PCA on the target domain and using the eigenvectors as basis to represent both 
the source and the target data ($\mathbf{PCA}_T$).

\vspace{2mm}
For all the methods the final classifier is a linear SVM with the $C$ parameter tuned by
two-fold cross-validation on the source over the range $\{0.001, 0.01, 0.1, 1.0, 10\}$.
Our \bata has three main parameters $(\alpha, \beta, d)$ that are also chosen by two-fold cross 
validation on the source. We remark that the target data are not annotated, thus tuning the 
parameters on the source is the only feasible option.
We searched for $\alpha,\beta$ in the same range indicated before for $C$. The parameter  
$d$ was tuned in $\{10, 20, \ldots, 100\}$ both for \bata and for the baselines 
$\mathbf{PCA}_T$, \textbf{TSL}, \textbf{TCA} and \textbf{SSTCA}.

We implemented the stochastic sub-gradient descent using a step size of $\eta = 0.1$ and a batch size 
of $\gamma=10$. The alternating optimization converges for less than 100 iterations and we can obtain the results 
for any of the source-target domain pairs of the Office+Caltech (excluding feature extraction) in 2 minutes using a modern desktop computer 
(2.8GHz cpu, 4Gb of ram, 1 core). With respect to the considered competing methods, % that focus only on the domain invariant representation, 
the training phase of JCSL is slower (e.g. 60 times slower with respect to SA and GFK),
but we remark that JCSL provides an optimized cross-domain classifier besides reducing the data distribution shift. 
The test phase runtime is comparable for all the considered approaches. In practical applications
domain adaptation models are usually learned offline, thus the training time is a minor issue.

\begin{figure*}[t]
 \centering
 \includegraphics[width=0.24\textwidth]{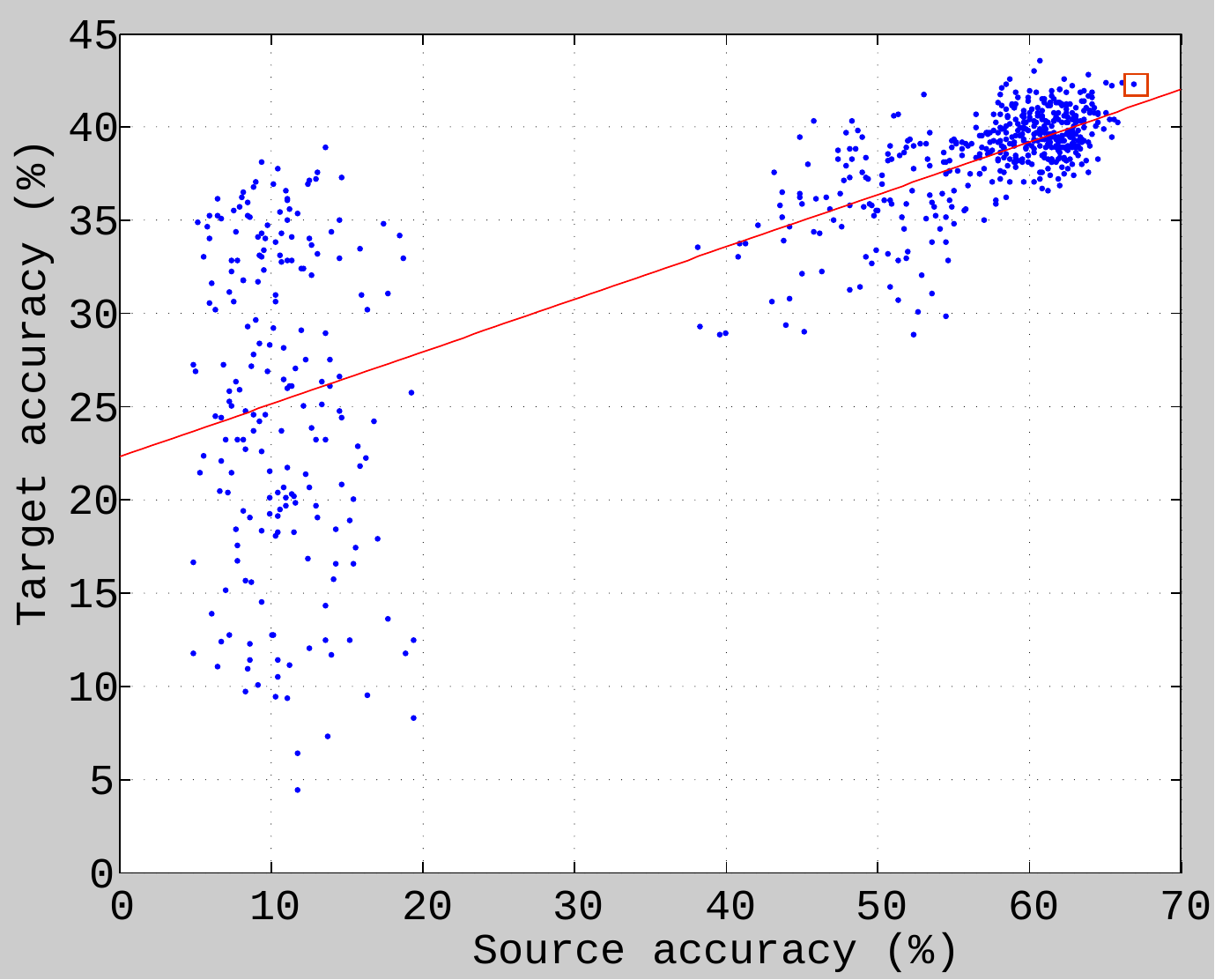}
 \includegraphics[width=0.24\textwidth]{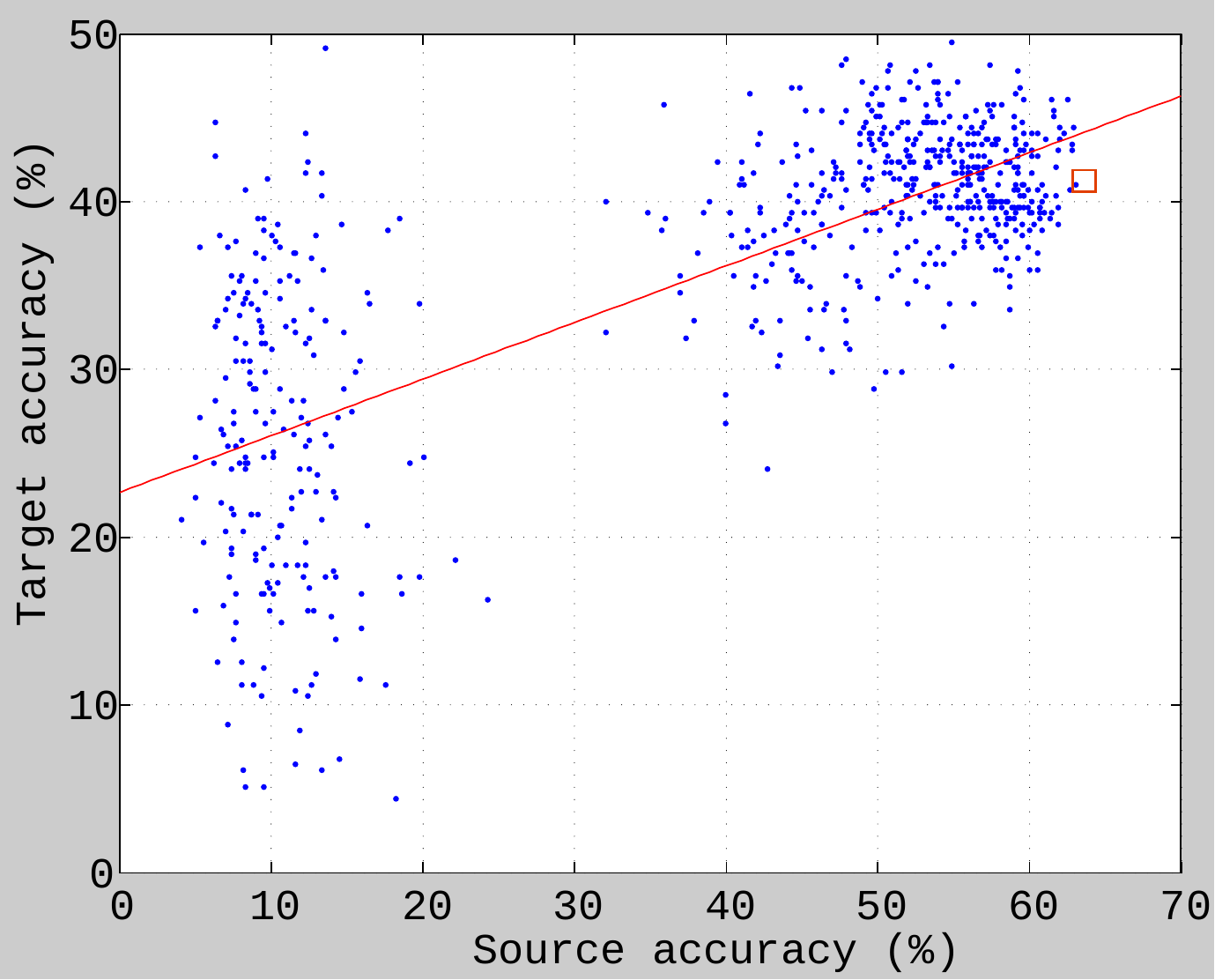}    
 \includegraphics[width=0.24\textwidth]{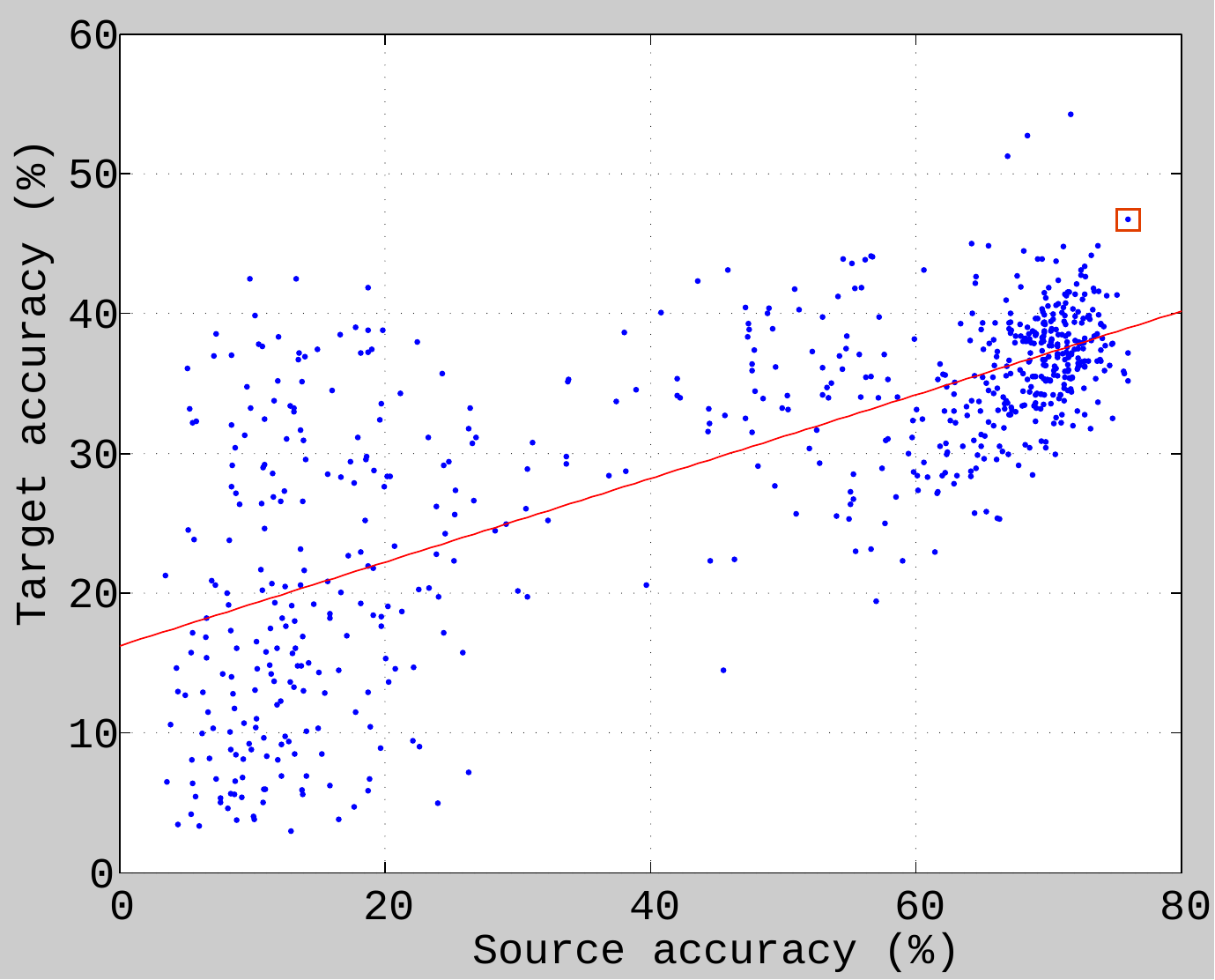}  
 \includegraphics[width=0.24\textwidth]{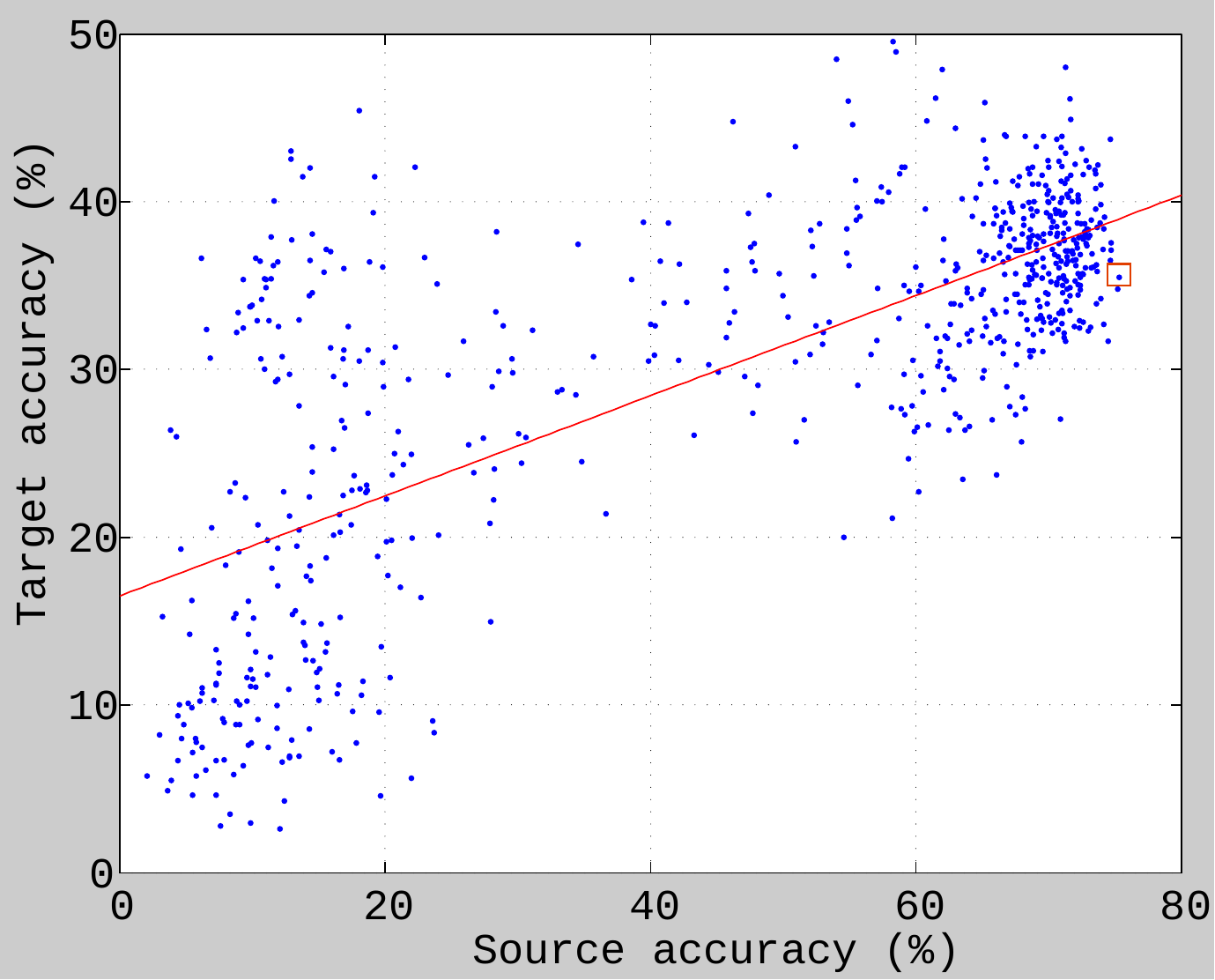}  
 \caption{Target accuracy vs source accuracy over domain adaptation problem 
(Amazon $\rightarrow$ Caltech), (Amazon $\rightarrow$ Webcam), (MNIST 
$\rightarrow$ USPS) and (USPS $\rightarrow$ MNIST) obtained by using \bata
and changing the parameters $\alpha,\beta$ and $d$. 
In all the cases the top right point cluster shows the high correlation between
the source and target accuracy. 
By comparing the x- and y- axis values of this
cluster it is also evident the source-to-target performance drop with respect 
to the source-to-source result in each experiment. 
The red square indicates the result selected by our method for the considered split.
The red line is obtained by least-square fitting and makes it evident the trend
in the results shared by all the source-target pairs.} \label{fig:st-accuracy}
 \end{figure*}
\begin{figure*}[t]
\centering
%\begin{tabular}{ c   c  c  c  c  c   }\hline 
%\dt{Bing(5)}{Caltech256} & 22.0 & 20.2 & 20.7 & 16.9 & \textbf{24.1} \\ 	
%\dt{Bing(10)}{Caltech256}& 25.0 & 21.6 & 20.6 &	23.2 & \textbf{27.4} \\ \hline 
%AVG. & 23.5 & 20.9 & 	20.6 & 	20.1 & \textbf{25.8} \\ \hline 
%\hline 
\begin{tabular} {c c}  
\includegraphics[width=0.35\textwidth]{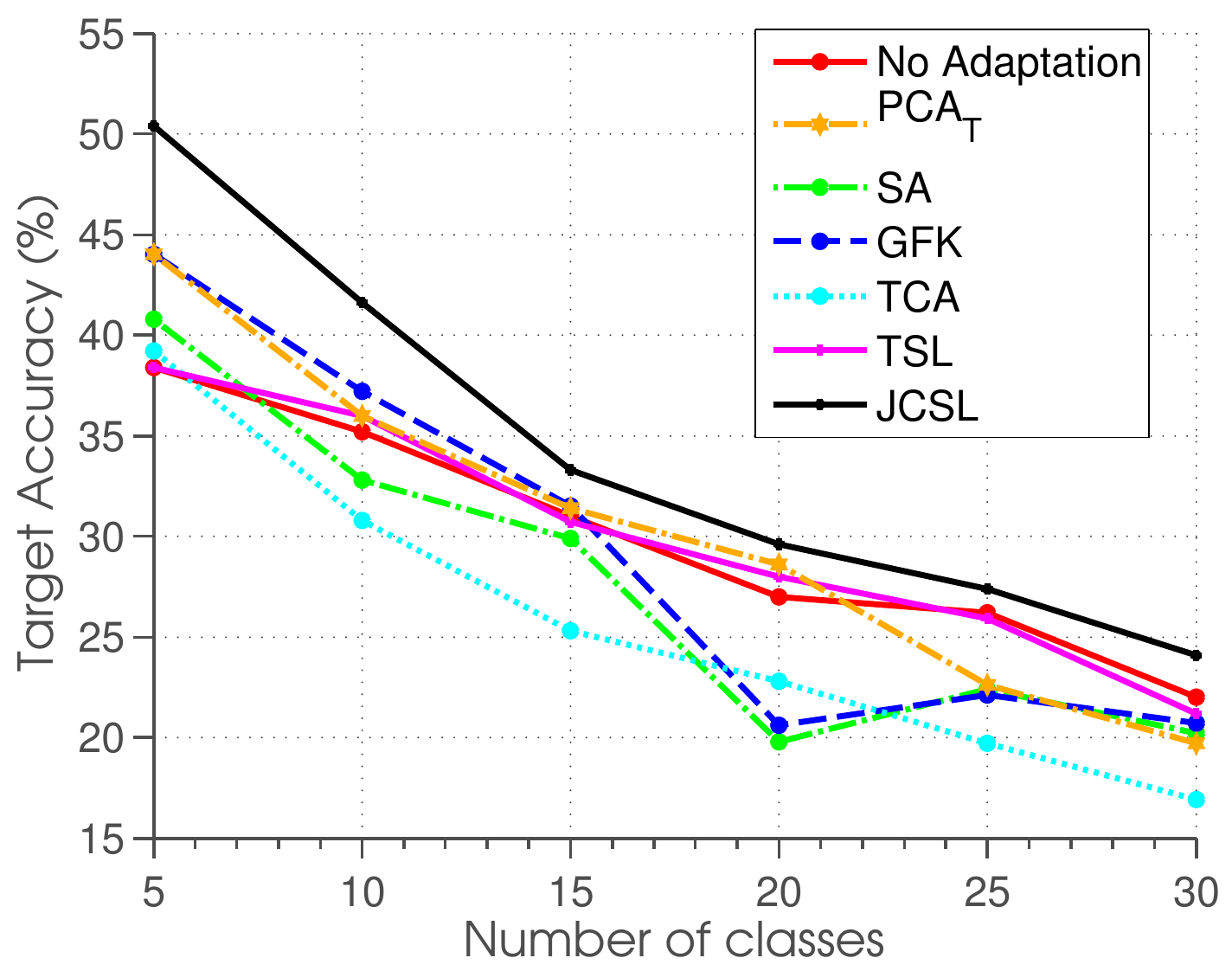} & \includegraphics[width=0.35\textwidth]{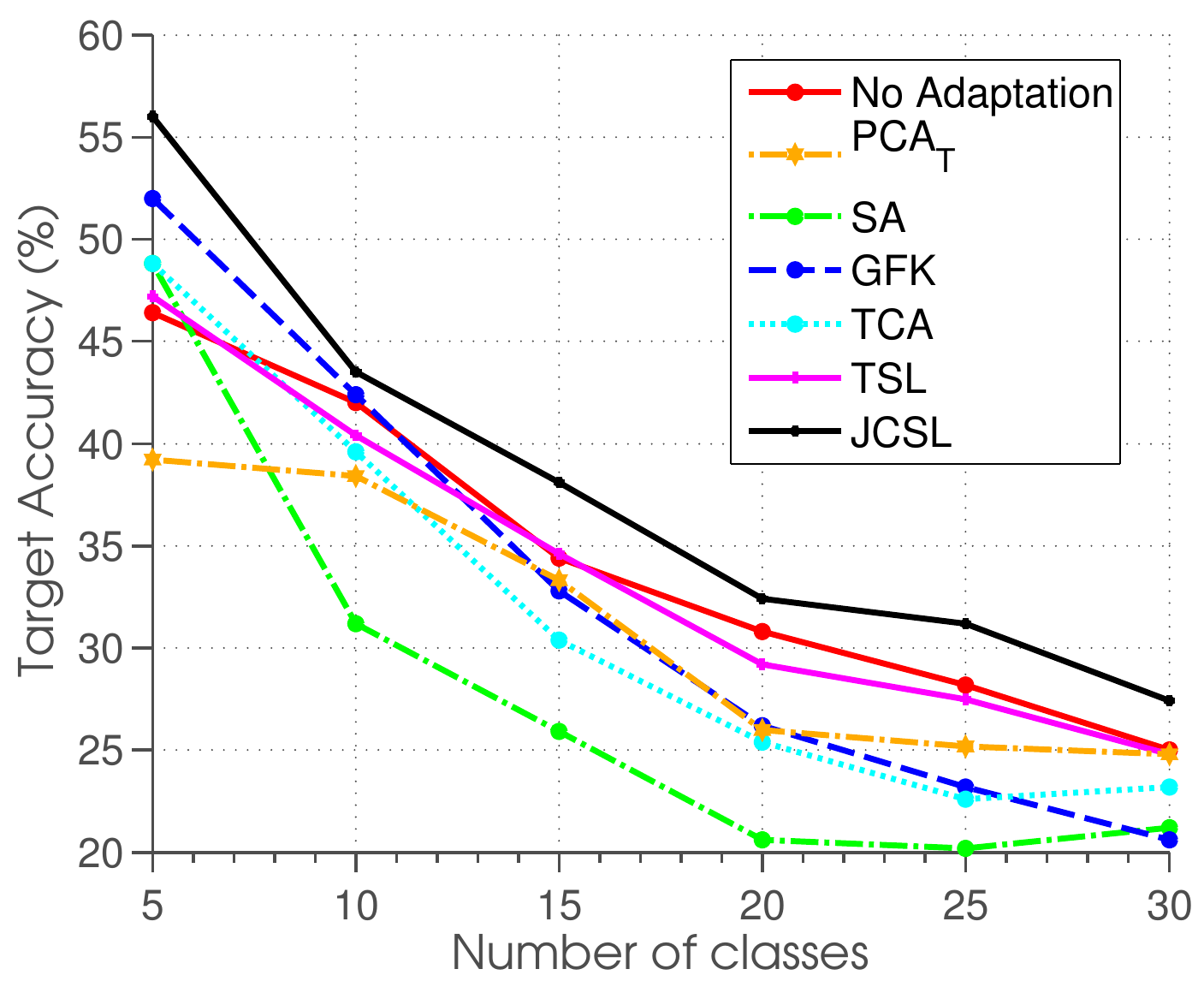}
\end{tabular} 
\caption{Experimental results on Bing+Caltech obtained when using 5 (left) and 10 (right) samples
per class in the source. SSTCA has shown similar or worse results than TCA, so we did not include it in this
evaluation to avoid further clutter in the plot.}
\label{fig:results_BC}
\vspace{-5mm}
\end{figure*}

\subsection{Results - Office+Caltech and MNIST+USPS}
\label{sec:results}

The obtained results over the Office+Caltech and MNIST+USPS datasets are presented in Table \ref{tbl:results}.
Overall \bata outperforms the considered baselines in 7 source-target pairs out of 14 and shows the
best average results over the two datasets. Thus, we can state that, minimizing a trade-off between source-target 
similarity and the source classification error pays off compared to only reducing the cross-domain representation divergence. Still SA
shows an advantage with respect to \bata in a few of the considered cases most probably because it can exploit
the discriminative LDA subspace. With respect to \bata, TCA and SSTCA seem to work particularly well when the domain shift 
is small (e.g. Amazon $\rightarrow$ Caltech, Dslr $\rightarrow$ Webcam). Interestingly \bata is 
the only method that consistently outperforms NA over MNIST+USPS.

\paragraph{Parameter analysis}
To better understand the performance of \bata we analyze how the target accuracy varies with respect to the source 
accuracy while changing the learning parameters $\alpha$, $\beta$ and $d$. 
The plots in Figure \ref{fig:st-accuracy} consider four domain adaptation problems,
namely (Amazon $\rightarrow$ Caltech), (Amazon $\rightarrow$ Webcam), 
(MNIST $\rightarrow$ USPS) and (USPS $\rightarrow$ MNIST)\footnote{Analogous results are obtained for all the remaining source-target pairs.}. 
All of them present 
two main clusters. On the left, when the source accuracy is low, the target
accuracy is uniformly distributed. This behavior mostly appears when 
$\beta$ is very small and $\alpha$ has a high value: this indicates that  
minimizing only $||V-T||_F^2$ does not guarantee stable results on the target task.
On the other hand, in the second cluster the source accuracy is highly correlated with the 
target accuracy. On average, for the points in this region, both the domain divergence term
and the misclassification loss obtain low values. The final \bata result 
with the optimal $(V^*,\mathbf{w}^*)$ appears always in this area and the dimensionality of the 
subspace $d$ seems to have only a moderate influence on the final
results. % indicating that the choice of this parameter is less relevant for JCSL than what is
%for the other subspace adaptive methods.
The red line reported on the plots is obtained by least-square fitting over the source and target
accuracies and presents an analogous trend for all the considered source-target pairs.
This is an indication that when domains are adaptable (negligible $\lambda$ in (\ref{bendavid})) 
our method is able to find a good source representation  as well as a classifier that 
generalizes to the target domain.

\paragraph{Measuring the domain shift}
For the same domain pairs considered above we also evaluate empirically the $\mathcal{H}\Delta\mathcal{H}$ divergence
measure defined in \cite{Ben-David2007}. This is obtained by learning a linear SVM that discriminates
between the source and target instances, respectively pseudo-labeled with $+1$ and $-1$. We separated
each domain into two halves and use them for training and test when learning a linear SVM model. 
A high final accuracy indicates high domain divergence. 
We perform this analysis by comparing the domain shift before and after the application of SA and JCSL,
according to their standard settings. 
SA presents a single step and learns one subspace representation $U$. JCSL exploits a one-vs-all 
procedure learning as many $V_y$ as the number of classes: each step involves all the data (no per 
class sample selection). The final domain shift for JCSL is the average over the obtained separate shift values.
%SA operates in a single step independently from
%the data multiclass nature and learns a single subspace representation $U$. JCSL exploits a one-vs-all 
%procedure learning as many $V_y$ as the number of classes. The $\mathcal{H}\Delta\mathcal{H}$ divergence
%obtained with SA is compared agains the average divergence of all the JCSL learned representations.
%Due to the considered one-vs-all multiple-class setting, our 
%\bata learns class-specific source subspaces $U_y$, so we calculate the average $\mathcal{H}\Delta\mathcal{H}$ divergence
%and compare it with the divergence obtained when using the original features and the SA target aligned subspace representation.   
The results in Table \ref{tbl:hdeltah} indicate that SA and \bata produce comparable results in 
terms of domain-shift reduction, suggesting that the main advantage of \bata comes from the learned classifier.
  
\begin{table}[t]
\caption{$\mathcal{H} \Delta \mathcal{H}$ analysis. Lower values indicate lower cross-domain distribution discrepancy.}
\centering
\scriptsize
\begin{tabular}{ | c | c | c | c | c |}\hline
Space 	& \dt{A}{C} & \dt{A}{W} & \dt{MNIST}{USPS} & \dt{USPS}{MNIST}\\ \hline
Original features   &  74.82  & 90.18	& 100.00 & 100.00\\ \hline
SA$_{\textsc{\tiny (LDA-PCA)}}$&  65.96  & 56.56	& 55.78 & 55.74\\ \hline
\bata   &  65.76  & 54.97	& 57.03 & 53.28\\ \hline
\end{tabular} 
\label{tbl:hdeltah}
\vspace{-3mm}
\end{table}

\subsection{Results - Bing+Caltech}
\label{sec:resultsBING}
Due to the way in which it was defined, Bing+Caltech can be considered as a much more 
challenging testbed for unsupervised domain adaptation compared to the other used 
datasets (see also Figure \ref{fig:datasets}). At the same time it also corresponds to
one of the most realistic scenarios where domain adaptation is needed: we have access 
to only a limited number of noisy labeled source images obtained from the web and we 
want to use them to classify over a curated collection of object images. For this problem 
exploiting at the best all the available information is crucial. Specifically, since
the source is not fully reliable, coding its discriminative information in the
representation (e.g. through LDA or PLS) may be misleading. On the other hand, using the subspace
of the non-noisy target data to guide the learning process can be much more beneficial.

As shown in Figure \ref{fig:results_BC}, \bata is the only method that consistently improves
over the non-adaptive approach independently from the number of considered classes. 
TSL is always equivalent to NA, while the other subspace methods,
although initially helpful for problems with few classes, lose their advantage over NA
when the number of classes increases. This behavior is almost equivalent when using both 5 and
10 source samples per class.

%In this setting we cannot even consider the covariate shift hypothesis as fully reliable 
%and with reference to the theoretical guarantee indicated by the bound in (\ref{}), we can 
%expect a higher $\lambda$ value than what 

%Deep analysis into the results obtained 
% from this dataset indicated that the source labeling function is quite 
% different from the target labeling function. This could also explain the poor 
% performance of almost all methods.

\subsection{Results - WiFi Localization}
\label{sec:resultsWIFI}
To demonstrate the generality of the proposed algorithm, we evaluate \bata also on non-visual data.
Since the WiFi vector dimensionality (100) is lower than the number of classes (247), we do not exploit LDA here
but we simply apply PCA to define the subspace dimensionality for both the source and target domains.
The results on the WiFi-localization task are reported in Table \ref{tbl:wifi} and show that 
domain adaptation is clearly beneficial.
\begin{table*}[t]
\caption{Classification accuracy obtained over WiFi localization dataset \cite{wifi}.
The \emph{full}-row contains the results over the whole target set. In the \emph{split}-row
we present the results obtained over 10 splits of the target, each containing 400 randomly 
extracted samples.}
\centering
\scriptsize
\begin{tabular}{| c | @{\:}c@{\:} | @{\:}c@{\:} | @{\:}c@{\:} | @{\:}c@{\:} | @{\:}c@{\:} | @{\:}c@{\:} |}\hline
\small
      &NA 		& SA$_{\textsc{\tiny (LDA-PCA)}}$		& GFK$_{\textsc{\tiny (LDA-PCA)}}$ 	& TCA 			& SSTCA 		& \bata\\ \hline
full  &16.6 	   	& 17.3  		& 17.3			& 19.0			& 18.5 			& 20.2 \\ \hline
splits&16.9 $\pm$ 2.1	& 17.6 $\pm$ 2.2	& 17.4 $\pm$ 2.1	& 19.2 $\pm$ 2.1	& 18.0 $\pm$ 2.2	& 20.5 $\pm$ 2.4\\ \hline
%full  &17.6	   	& 17.6  		& 17.1			& 18.5			& 18.4 			& ??? \\ \hline
%splits&18.5 $\pm$ 1.4   & 18.5 $\pm$ 1.4 	& 17.7 $\pm$ 1.2   	& 18.6 $\pm$ 2.2 	& 18.7 $\pm$ 2.3	& ??? $\pm$ ??    \\ \hline
\end{tabular}
\label{tbl:wifi}
\vspace{-3mm}
\end{table*}
%It should be noted that WiFi localization is relatively a difficult classification task. 
%However, all subspace based domain adaptation methods managed to obtain better results 
%than no adaptation indicating that domain adaptation is useful in this task. 
%It is interesting to see that unsupervised PCA based methods (SA, GFK) obtain 
%similar results while TCA method manages to obtain better results than SA method 
%and the GFK method. Interestingly, \bata method outperforms TCA by 1.2\% indicating 
%the general applicability of \bata on non-computer vision datasets.
TCA and SSTCA are the state of the art linear methods on the WiFi dataset and they confirm their 
value even in the considered classification setting by outperforming SA and GFK. Still JCSL presents the
best results. The obtained classification accuracy confirms the value of our method over the 
other subspace-based techniques.

\section{Conclusions}
\label{conclusions}

Motivated by the theoretical results of Ben-David et al. \cite{Ben-David2007}, in this
paper we proposed to integrate the learning process of the source prediction function 
with the optimization of the invariant subspace for unsupervised domain adaptation.
Specifically, \bata learns a representation that minimizes the divergence 
between the source subspace and the target subspace, while optimizing the classification model.
Extensive experimental results have shown that, by taking advantage of the described principled
combination and without the need of passing through the evaluation of the data distributions, 
\bata outperform other subspace domain adaptation methods that focus only on the 
representation part. 

Recently several works have demonstrated that Convolutional Neural Network classifiers 
are robust to domain shift \cite{DonahueJVHZTD13,Oquab14}. Reasoning at high level we can identify 
the cause of such a robustness on the same idea at the basis of \bata: deep architectures learn jointly
a discriminative representation \emph{and} the prediction function. The highly non-linear
transformation of the original data coded into the CNN activation values can also be used as 
input data descriptors for \bata with the aim of obtaining a combined effect. As future 
work we plan to evaluate principled ways to find automatically the best 
subspace dimensionality $d$ using low-rank optimization methods.

% 
% 
% 
% 
% As future work we plan to investigate on  principled ways to find automatically the best 
% subspace dimensionality $d$ using low-rank optimization methods.
% 
% presented a new subspace-based domain adaptation method named \bata
% that focuses on the difficult unsupervised DA setting. 
% Motivated by the theoretical results of Ben-David et al. \cite{Ben-David2007} our 
% algorithm jointly seek a good data representation and a prediction model. 
% 
% Specifically, \bata learns a representation that minimizes the Bregman matrix divergence between the source subspace
% and the target subspace while optimizing the classification model.
% Extensive experimental results show that our approach outperform several other
% subspace domain adaptation methods taking advantage of learning an effective
% cross-domain classifier.
% 
% As future work we plan to investigate on  principled ways to find automatically the best 
% subspace dimensionality $d$ using low-rank optimization methods.
% 
% It is necessary to explain that the learning part is here injected into a subspace 
% learning method but the concept is general and can be combined also with MMD enlarging
% other existing domain adaptation methods with the aim to let them optimize the indicated
% theoretical bound.

\section*{Acknowledgments}
The authors acknowledge the support of the FP7 EC project AXES and of the FP7
ERC Starting Grant 240530 COGNIMUND.

\end{document}